\documentclass[journal]{IEEEtran}

\usepackage{times}
\usepackage{epsfig}
\usepackage{graphicx}
\usepackage{amsmath}
\usepackage{amssymb}
\usepackage{tabularx}
\usepackage{multirow}
\usepackage{subfigure}
\usepackage{array}
\usepackage[colorlinks,pdfstartview=FitH]{hyperref}

\def\First#1{\textbf{\color{red}{#1}}}
\def\Second#1{\textit{\textbf{\color{blue}{#1}}}}

\newcolumntype{Y}{>{\centering\arraybackslash}X}

\usepackage{ifpdf}
\ifpdf   
  \graphicspath{{./pdf/}{./pic/}}
   \DeclareGraphicsExtensions{.pdf,.jpeg,.png,.jpg}
\else    
  \graphicspath{{./eps/}}
  \DeclareGraphicsExtensions{.eps,.ps}
\fi

\ifCLASSINFOpdf
\else
\fi

\begin{document}

\title{Structured Inhomogeneous Density Map Learning for Crowd Counting}

\author{{Hanhui Li, Xiangjian He, Hefeng Wu, Saeed Amirgholipour Kasmani, Ruomei Wang, Xiaonan Luo, Liang Lin}
\thanks{H. Li, R. Wang and L. Lin are with the School of Data and Computer Science, Sun Yat-sen University, Guangzhou 510006, China.}%
\thanks{X. He and S. Kasmani are with the University of Technology Sydney, Sydney, Australia.}%
\thanks{H. Wu is with the School of Information Science and Technology, Guangdong University of Foreign Studies, Guangzhou 510006, China.}%
\thanks{X. Luo is with the School of Computer Science and Information Security, Guilin University of Electronic Technology, Guilin 541004, China.}%
}

\maketitle

\begin{abstract}
In this paper, we aim at tackling the problem of crowd counting in extremely high-density scenes, which contain hundreds, or even thousands of people. We begin by a comprehensive analysis of the most widely used density map-based methods, and demonstrate how easily existing methods are affected by the inhomogeneous density distribution problem, e.g., causing them to be sensitive to outliers, or be hard to optimized. We then present an extremely simple solution to the inhomogeneous density distribution problem, which can be intuitively summarized as extending the density map from 2D to 3D, with the extra dimension implicitly indicating the density level. Such solution can be implemented by a single Density-Aware Network, which is not only easy to train, but also can achieve the state-of-art performance on various challenging datasets.

\end{abstract}

\section{Introduction}
Counting people in crowded scenes is a highly challenging and important problem in  computer vision and video surveillance. According to the statistics\footnote{\url{https://en.wikipedia.org/wiki/List_of_human_stampedes}}, there are at least $9$ serious human stampedes happened in 2017, causing at least $70$ people to die, and thousands of people were injured. Such tragedy can be prevented if we can estimate the crowd density and take preventive measures in time.

\begin{figure}[!t]
\centering
\includegraphics[width = 1\columnwidth]{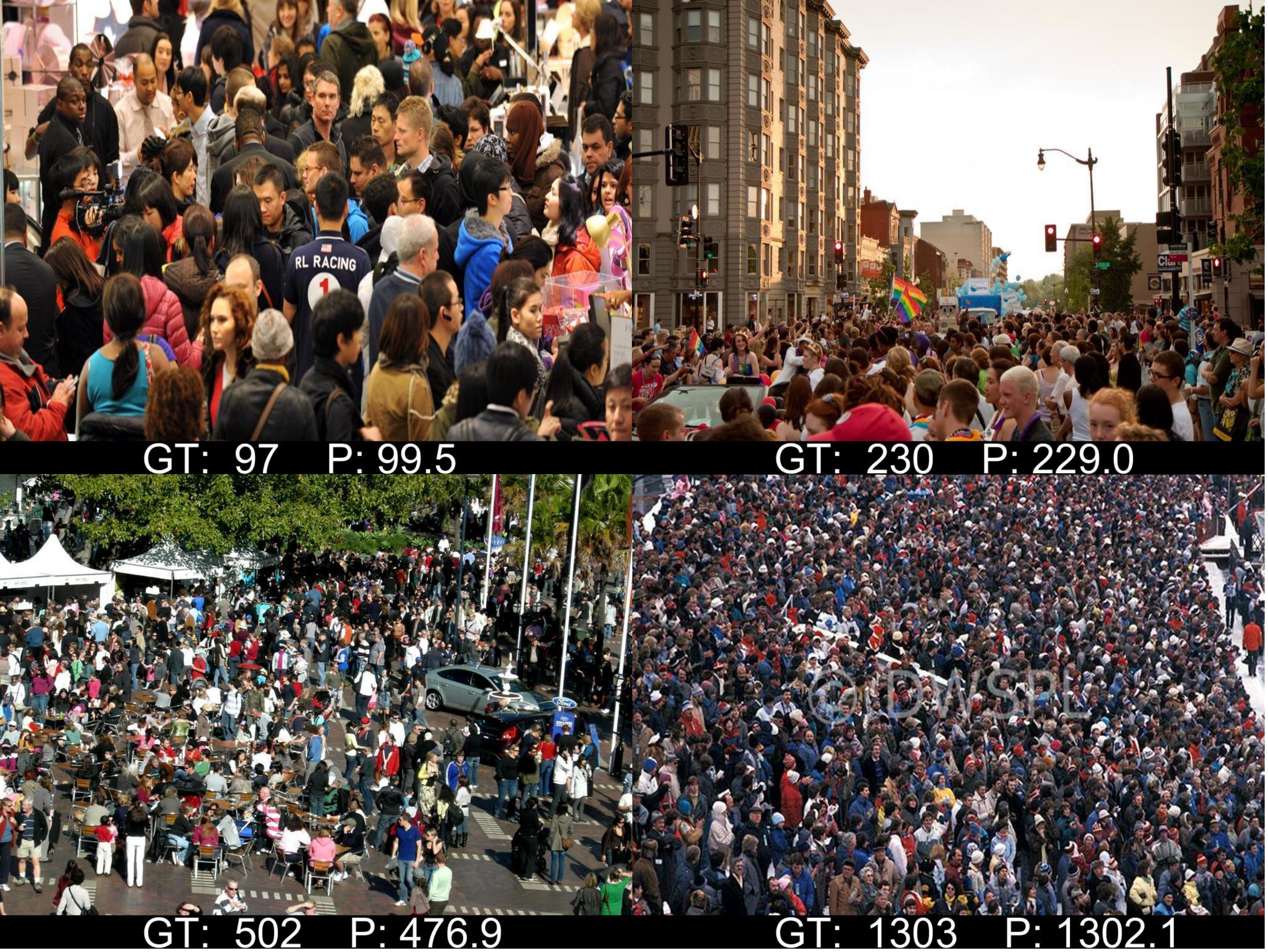}
\caption{Test images from the ShanghaiTech A \cite{ShanghaiTech} dataset. The goal of this paper is to calculate the number of people in images with inhomogeneous density distribution. The ground-truth (GT) and predicted counts (P) of our method are marked on the bottom of each image.}
\label{fig:cover}
\end{figure}

Precisely detecting or recognizing each person is intractable in crowed scenes due to the heavy occlusions and cluttered visual patterns, as shown in Figure \ref{fig:cover}. Therefore, most existing methods \cite{ShanghaiTech,BodyCNN2017,MSCNN,convLSTM,CP-CNN2017,SwitchCNNCVPR2017,Cascaded-MTL2017,boostedCNN,HyderNet,WorldExpo} choose to estimate the density map, of which the integral is equal to the number of people in a given image. With the recent breakthrough of deep learning in visual related tasks, most cutting-edge counting methods are based on the Convolutional Neural Network (CNN), e.g., MCNN \cite{ShanghaiTech} and Hydra \cite{HyderNet}.

However, our study reveals an interesting phenomenon that, in the current all ``going deeper'' era, the network architectures in most counting methods are rather shallow. For example, MCNN only uses $5$ convolutional layers with a limited number of filters. The same situation exists even in many multi-subnet networks \cite{SwitchCNNCVPR2017,Cascaded-MTL2017,CP-CNN2017}, e.g., the core component of Switch-CNN \cite{SwitchCNNCVPR2017} for generating density maps is exactly the same as in MCNN. Yet this is not simply because of hardware limitation, e.g., the auxiliary component in Switch-CNN, a density-level classifier, uses the very deep VGG-16 \cite{VGG-16} structure.

In fact, learning a Deep Neural Network (DNN) to estimate density map is not so straightforward and even frustrating. In both \cite{HyderNet} and \cite{boostedCNN}, the difficulty of optimizing their own DNN has been mentioned. Besides, the earlier research in \cite{deepRegression} also points out that learning a DNN with the $\ell_2$-loss, which is the major objective in most density map based methods, is vulnerable to outliers.

Based on the above considerations, we conduct in-depth study to dig out the main reason behind the disadvantages of learning DNNs for density map estimation. After delving into the learning process of existing methods, we attribute the main reason to the \emph{inhomogeneous density distribution} in crowd images. By inhomogeneous density distribution, we refer to the various intra-image and inter-image density levels. As we will see in the later sections, it will not only introduce outliers, but also result in other problems, such as the dying ReLU phenomenon, in the optimization process.

Fortunately, with the comprehensive analysis of the inhomogeneous density distribution problem, we are able to find an intuitive and natural solution for it: we simply transform our learning labels from 2D density maps to structured (3D) density maps. Such solution can reduce the effects of outliers sufficiently, and with a few modifications on the VGG-16 network, we successfully train a single Density-Aware Network (DAN), which obtains comparable performance with the much more complicated multi-subnet networks.

In summary, this paper focuses on the inhomogeneous density distribution problem in the high-density crowd counting task. The major contributions of this paper are dual:
(a) We present the detailed analysis of how the inhomogeneous density distribution affects the density map based methods. We also point out the reasons of several abnormal phenomena in the optimization process, and provide solutions for them. We hope this can motivate other researchers to handle similar problems in their optimization process.
(b) We propose the structured inhomogeneous density learning method as a practical solution for crowd counting, which is implemented by our DAN model. Extensive experiments on several datasets show that our simple method can obtain the state-of-the-art performance.

\section{Related Work}
Earlier research on crowd counting mainly aims at low density scenes, and can be roughly divided into detection-based methods \cite{DPM,countlocalization} and regression-based methods \cite{FeatureMining,TransferCounting2013,Cumulative2013,BayesianRegression12,InterCount2014}. In recent years, as high-density datasets \cite{UCF,WorldExpo,ShanghaiTech} are built, density map-based methods, which is firstly introduced in \cite{Lempitsky}, have gained more attention. Due to the page limitation, we focus on density map-based methods.

Most of the state-of-the-art density map-based methods are implemented using CNNs and its variants. A comprehensive survey of CNN-based counting models can be found in \cite{surveyPRL}. Based on the network architecture, we can divide existing methods into (a) single-column networks, (b) multi-column networks and (c) multi-subnet networks, as shown in Figure \ref{fig:smap}:

\textbf{Single-Column Networks}: we consider that networks \cite{HyderNet,convLSTM} with the straight structure similar to the Fully Convolutional Networks \cite{FCN} as the single-column networks. A typical example of this kind of networks is the Counting CNN (CCNN, \cite{HyderNet}), which is simply trained by minimizing the $\ell_2$-loss between its outputs and the ground-truth density maps. In \cite{convLSTM}, the CNN is combined with the LSTM to capture spatial and temporal dependencies. The advantage of this kind of networks is the simplicity of their architectures.

\textbf{Multi-Column Networks}: multi-column networks, or multi-scale networks are introduced to tackle the scale variance and distorted perspective map problem \cite{ShanghaiTech,HyderNet,boostedCNN,MSCNN}. This type of networks usually use multiple columns with the same functions to capture features at different resolutions. For example, MCNN \cite{ShanghaiTech} uses three columns of filters of different sizes; Hydra \cite{HyderNet} uses a pyramid of input patches; and in \cite{MSCNN}, a multi-scale blob with different kernel sizes is introduced. Multi-column networks are usually more robust than the single column networks, since they can integrate features with different resolutions. However, as we will point out later, multi-column networks actually are global models, they will still be affected by the outliers in the training data.

\textbf{Multi-Subnet Networks}: multi-subnet networks \cite{CP-CNN2017,BodyCNN2017,SwitchCNNCVPR2017,Cascaded-MTL2017,boostedCNN}, or multi-task networks are networks consisting of multiple subnets, and the main difference between multi-subnet networks and multi-column networks is that each subnet in multi-subnet networks has its own objective function. For example, Switch-CNN \cite{SwitchCNNCVPR2017} is composed of three local regressors and one density-level classifier. The three regressors are trained to generate the their own density maps, while the density-level classifier is trained to select the optimal density map. A more complicate example is the CP-CNN \cite{CP-CNN2017} model, which has a global-context classifier, a local-context classifier, a density map regressor, and a fusion network to combine these subnets together. Indeed, among the three kinds of networks, multi-subnet networks can obtain the most impressive performance. However, learning a multi-subnet network is difficult, since it requires individually tuning for its each component.

\begin{figure}[!t]
\centering
\includegraphics[width = 1\columnwidth]{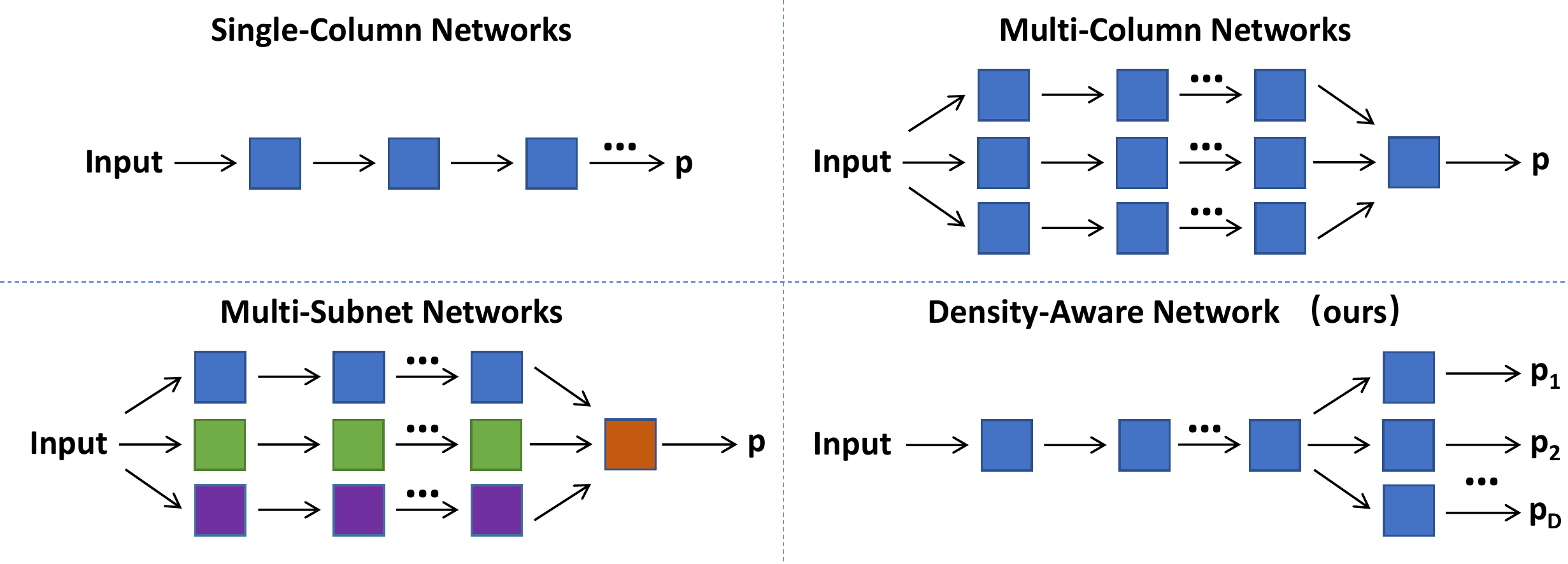}
\caption{Comparison of the abstracted structure of the proposed Density-Aware Network (DAN) with existing density map-based networks. Here $p$ denotes density map. The main difference between DAN and other methods is that DAN can predict the structured density maps.}
\end{figure}

\section{Crowd Counting with Density Map}
In this section, we first briefly overview density map based counting methods in Section \ref{sec:formulation}, and then present the detailed analysis of how the inhomogeneous density distribution problem affects existing methods in Section \ref{sec:density}, so that we can solve it accordingly in our method.

\subsection{Problem Formulation}
\label{sec:formulation}
Given an image $I$ belonging to the image domain $\mathcal{I}$, the primary goal of this paper is to learn a function $f$ to estimate the number of people in $I$, i.e., $f: \mathcal{I} \to \{0,{\mathbb{R}^ + }\}$. To this end, we are given a training set consisting of $N$ images and their corresponding ground-truth head based annotations, i.e., $\mathcal{T} = \{(I_1,A_1),(I_2,A_2),...,(I_N,A_N)\}$, where $A_n = (a^{ij}_n) \in \{0,1\}^{H_n \times W_n}$, $n = 1,...,N$, $H_n$ and $W_n$ denote the height and width of the image, respectively, and $a^{ij}_n = 1$ if the center of a target appears at position $(i,j)$, otherwise $a^{ij}_n = 0$, $i = 1,...H_n$, $j = 1,..,W_n$. Yet the information provided only by the count is limited, therefore we are also interested in inferring the density maps of the images. Specifically, given the annotation $A_n$, the corresponding ground-truth density map ${\bf{g}}_n$ is generated as follows:
\begin{equation}
{{\bf{g}}_n}{\rm{ = }}\sum\limits_{{(x,y)} \in {P_n}} {G(x,y,{\sigma _g})},
\label{eq:densityMap}
\end{equation}
where $P_n$ denotes the set of non-zero points in $A_n$, and $G(x,y,{\sigma _g})$ denotes a 2D normalized Gaussian function centered at point $(x,y)$ with smoothing factor $\sigma_g$\footnote{For the compactness of expression, we have omitted the spatial transformation process of $P_n$.}. The size of ${\bf{g}}_n$ depends on the specific feature representation, and we consider it as $sH_n \times sW_n$, where $s$ is the scaling factor. Calculating the count $c^{{g}}_n$ based on the density map ${\bf{g}}_n$ can be done easily by summing over elements in ${\bf{g}}_n$, i.e., $c^{{g}}_n = \sum_{i,j} {{g^{ij}_n}}$. In this way, most density map based methods can be summarized as learning a regression model by minimizing the following loss function:

\begin{equation}
L = \frac{1}{N} \sum\limits_{n = 1}^N {{{\left\| {f({I_n}) - {{\bf{g}}_n}} \right\|}_2^2} + {\cal R}(f)},
\label{eq:regression}
\end{equation}
where ${|| \cdot ||}_2$ denotes the $\ell_2$-loss, and ${\cal R}(f)$ can be any differential regularizer. Like in other deep learning models, gradient descent or its variants can be adopted to minimize $L$.

\label{sec:DAN}
\begin{figure*}[!tbh]
\centering
\includegraphics[width = 1\textwidth]{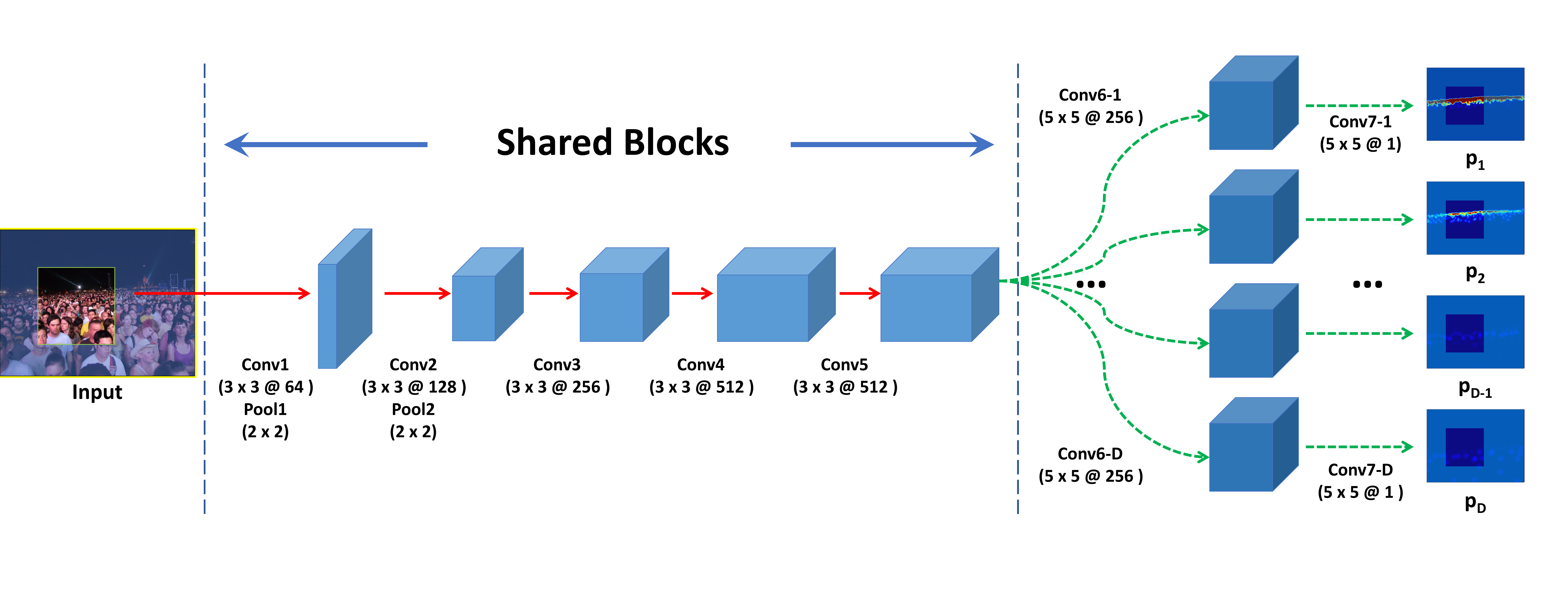}
\caption{Architecture of the proposed DAN. The shared blocks part is composed of the \emph{Conv1} to \emph{Conv5} blocks in VGG-16 \cite{VGG-16}. $D$ branches for generating the structured density maps are attached to the end of the shared blocks.}
\label{fig:architecture}
\end{figure*}

\subsection{The Inhomogeneous Density Distribution}
\label{sec:density}

From the experimental results in previous research \cite{ShanghaiTech,boostedCNN,SwitchCNNCVPR2017,CP-CNN2017}, we observe three interesting phenomena: (a) Compared with global model, local model shows superior performance. This is observed from the fact that the three columns of CNNs in Switch-CNN \cite{SwitchCNNCVPR2017} and those in MCNN \cite{ShanghaiTech} are the very same, and the major difference between these two networks is Switch-CNN discriminatingly selects one column for predicting, while MCNN merges all three columns into a global regressor. (b) Compared with shallow networks, quite surprisingly, DNNs perform poorly, especially on datasets with various density levels \cite{boostedCNN}. (c) The mean absolute error (MAE) of current methods \cite{ShanghaiTech,CP-CNN2017} is proportional to the relative density level. If we divide images into groups based on their density levels, i.e., extremely low to extremely high, and consider the group with medium density level as the baseline, then groups far from the baseline suffer higher MAE than those near the baseline. All these phenomena indicate that blindly increasing the capacity of networks cannot help to solve our task. In order to increase the robustness of our solution, we must dig out the main reason behind the above phenomena, which, in fact, is the \emph{inhomogeneous density distribution} in our images.

To understand how the inhomogeneous density distribution problem affects the performance of current methods, note that in the loss function $L$ defined in Eq.(\ref{eq:regression}), the major objective is to minimize the $\ell_2$-loss term, which is easily affected by outliers. Without loss of generality, we take MCNN as an example. For the convenience of discussion, we only consider its last layer, which contains a filter group of size $1 \times 1 \times 30$, and ReLU is used as the activation function. We assume the input of the last layer is noise-free. Let ${\bf{w}} = [{w^1},{w^2},...,{w^k},...,{w^{30}}]$ and $b$ denote the weights and bias of the filters, ${\bf{z}}_n \in {\{0,{\mathbb{R}}^+\}}^{sH_n \times sW_n \times 30}$ denote the input of the last convolutional layer, and ${\bf{r}}_n$ be short for the residual that ${\bf{r}}_n = f(I_n) - {\bf{g}}_n$, based on the chain rule we can calculate the partial derivative of $\ell_2$-loss term with respect to $w^k$ and $b$ as
\begin{equation}
\left\{ \begin{array}{l}
\frac{{\partial ||{{\bf{r}}_n}|{|^2_2}}}{{\partial {w_k}}} = 2\sum\limits_{i,j} {[r_n^{ij} + g_n^{ij} > 0} ] \cdot r_n^{ij} \cdot z_n^{ijk}\\
\frac{{\partial ||{{\bf{r}}_n}|{|^2_2}}}{{\partial b}} = 2\sum\limits_{i,j} {[{r^{ij}_n} + g_n^{ij} > 0} ] \cdot r_n^{ij}
\end{array} \right.,
\label{eq:derivative}
\end{equation}
where $[\cdot]$ is the indicator function. From Eq.(3) we can see that, for all ${r_n^{ij} \ge -g_n^{ij}}$, the gradient is dominated by $r_n^{ij}$. Therefore, if outliers exist, e.g., ${\bf{g}}_n$ is affected by additive noise ${\bf{e}}_n = (e^{ij}_n) \in {\mathbb{R}}^{sH_n \times sW_n}$, then $r_n^{ij}$ is equally affected by the bias $e_n^{ij}$, and consequently disrupts the optimization process.

But where do the outliers come from? Remember that $\bf{g}$ is generated by convolving the head annotation map $A$ with the 2D Gaussian kernel (Eq.(\ref{eq:densityMap})), of which the essential idea is, the sum of density values in the area of one head is equal to $1$. Therefore, in ideal situation, we should generate the high and compact Gaussian response for people far from the camera (high-density area), while the low and flat response for those near the camera (low-density area). However, due to the lack of annotation, current methods have to use a ``moderate" Gaussian template to generate the density map. In this way, for high-density / low-density areas, the density values of the real training template are lower / higher than those of the ideal one, and consequently the learned model will underestimate / overestimate the counts, which is exactly in accordance with the experimental results in \cite{CP-CNN2017}.

With the inhomogeneous density distribution, the aforementioned phenomena can be explained well now: (a) Since Switch-CNN selects the training images for each of its regressors individually, it actually removes the outliers in a certain extent. On the other hand, the geometry-adaptive kernel in MCNN does help to reduce the bias, but the outliers remain in the training process. (b) With deeper architectures, networks are more likely to over-fit the outliers, degrading their generalization ability. (c) The real training template is quite different from the ideal response in areas with extreme density levels, therefore the trained models are more affected by the outliers in these areas.

Besides, by analyzing the gradient of the $\ell_2$-loss term, we can also infer the following problems will occur if we train a deep neural network to minimize the $\ell_2$-loss without deliberate consideration:

\textbf{Dying ReLU}: For any $\frac{{\partial ||{{\bf{r}}_n}|{|^2_2}}}{{\partial {w_k}}} > 0$, i.e., when the network overestimates the density, if the value of $\frac{{\partial ||{{\bf{r}}_n}|{|^2_2}}}{{\partial {w_k}}}$ or the learning rate for gradient descent is too large, then $w_k$ will decrease dramatically, causing the activation value $\sum\nolimits_k {{w_k}{z_k}}  < 0$ for $z_k$ in a wide range of values. Since the gradient can only pass through a ReLU when the activation value is larger than $0$, it is hard to update $w_k$ in this case, and consequently the output of the ReLU will always be $0$.

\textbf{Exploding Gradients}: Similar to Eq.(\ref{eq:derivative}), we have $\frac{{\partial ||{{\bf{r}}_n}|{|^2_2}}}{{\partial z_n^{ijk}}} = 2[r_n^{ij} + g_n^{ij} > 0] \cdot r_n^{ij} \cdot {w^k}$, so if the value of $r_n^{ij} \cdot {w^k}$ is too large, e.g., $r_n^{ij} \cdot {w^k} > 1$, it may cause the gradient to explode. In fact, Daniel et al. \cite {HyderNet} mention that their CCNN cannot converge with the Xavier Initialization \cite{Xavier}. Although they didn't explain the reason behind it, we can see clearly that it is caused by the $r_n^{ij} \cdot {w^k}$ term.

\textbf{Saddle Points}: The gradient is dominated by ${\bf{r}}_n$, and all $r^{ij}_n$ in ${\bf{r}}_n$ contribute equally. Hence, if there are a few $r^{ij}_n$ with unbounded, extremely high values, they may cancel out the partial derivative of the other elements in ${\bf{r}}_n$, namely reaching the critical points. Furthermore, \cite{saddle} points out that if the critical points are with $e^{ij}_n$ much larger than the global minimum, then they are exponentially likely to be the saddle points and will slow than the learning process\footnote{For the relationship between critical points and saddle points, interested readers can refer to \cite{saddle} for more details}.

\section{Density-Aware Network}
Based on the aforementioned analysis of the inhomogeneous density distribution problem, we can see that the ways of tackling it are quite straightforward: we can either (a) minimizing the difference between ideal templates and real training templates, or (b) reduce the effects of outliers, or, as we will show in Section \ref{sec:structureLearning}, we can even combine and accomplish both goals simultaneously within a unified structured learning framework, and implement the framework easily via the proposed Density-Aware Network (DAN) in Section \ref{sec:DAN}.

\subsection{Separating Inhomogeneous Density Distribution as Structured Learning}
\label{sec:structureLearning}
Our solution for the inhomogeneous density distribution problem is intuitive and natural: we extend the 2D density maps to the structured density maps (3D), of which the last dimension implicitly indicates the density levels, and we use an individual Gaussian kernel on each density level.

Formally, our goal now is to learn a structured regressor $f: \mathcal{I} \to \{0, {\mathbb{R}^ + }\}^{H \times W \times D}$, where $D$ denotes the number of predefined density levels. In order to do so, we need to convert the original head annotation $a_n^{ij}$ into the new structured label ${\bf{v}}_n^{ij} = [v_n^{ij1},...,v_n^{ijD}]$, $v_n^{ijd} \in \{0,\mathbb{R}^{+}\}$, $d = 1,...,D$. Here we propose a heuristic soft mapping rule to determine the value of $v_n^{ijd}$ for $a_n^{ij} = 1$:

\begin{equation}
v_n^{ijd} = \frac{{1}}{{{\rm{Z}}_n^{ij}}} \cdot {e^{ - \frac{{{{(d - d_n^{ij*})}^2}}}{{2{\sigma _v}}}}},
\label{eq:vlabel}
\end{equation}
where ${{\rm{Z}}_n^{ij}}$ is the normalization term to make sure $\sum_d {v_n^{ijd}}  = 1$, $\sigma _v$ is the empirical smoothing parameter, and $d_n^{ij*}$ denotes the index of the optimal density level for $a_n^{ij}$. To determine $d_n^{ij*}$, notice that the average distances between annotated points in the high-density areas are usually much smaller than those in the low-density areas, therefore, we simply define a strictly increasing thresholds $S_1,S_2,...,S_D$, and calculate the average distance from point $(i,j)$ to its Top-5 nearest annotated points, and consider $d_n^{ij*}$ as the index of the minimum threshold that large than or equal to the average distance. Similar to Eq.(\ref{eq:densityMap}), we generate the structured density map on level $d$, ${\bf{g}}_n^{d}$, by convolving ${\bf{v}}_n^d$ with a normalized Gaussian filter $G_d$ with the empirical smoothing parameter $\sigma_d$.

\begin{figure}[!t]
\centering
\includegraphics[width = 1\columnwidth]{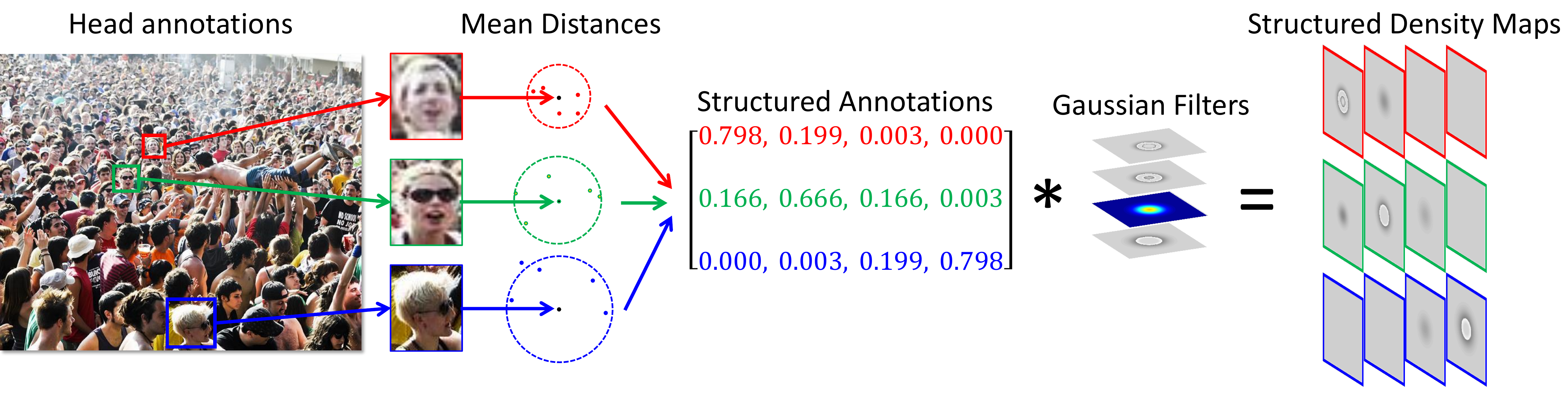}
\caption{Demonstration of the generation of structured density maps. Given an annotated point, we first calculate its mean distance to its Top-K neighbors, and then use the proposed soft mapping method to convert it to the structured annotation of multiple levels. Different colors denote different density levels in this figure. Lastly, we convolve each level of the structured annotation with a Gaussian filter to generate the structured density map.}
\label{fig:smap}
\end{figure}

\textbf{Remark}: Figure \ref{fig:smap} shows a toy example of the generating process of the structured density maps. The benefits of learning such structured density maps are trial: (a) Compared with the single Gaussian template solution, apparently multiple templates are less biased, and hence are more suitable to model the inhomogeneous density distribution; (b) Each level learns to generate its own density maps resembling the local model, thus can help to reduce the effects of outliers. (c) The outputs of all levels are generated simultaneously, which naturally serves as the regularizer, e.g., it is unlikely for a position to have high response values on both the high-density and the low-density levels.

\subsection{Robust Regression}
\begin{figure}[!t]
\centering
\includegraphics[width = 1\columnwidth]{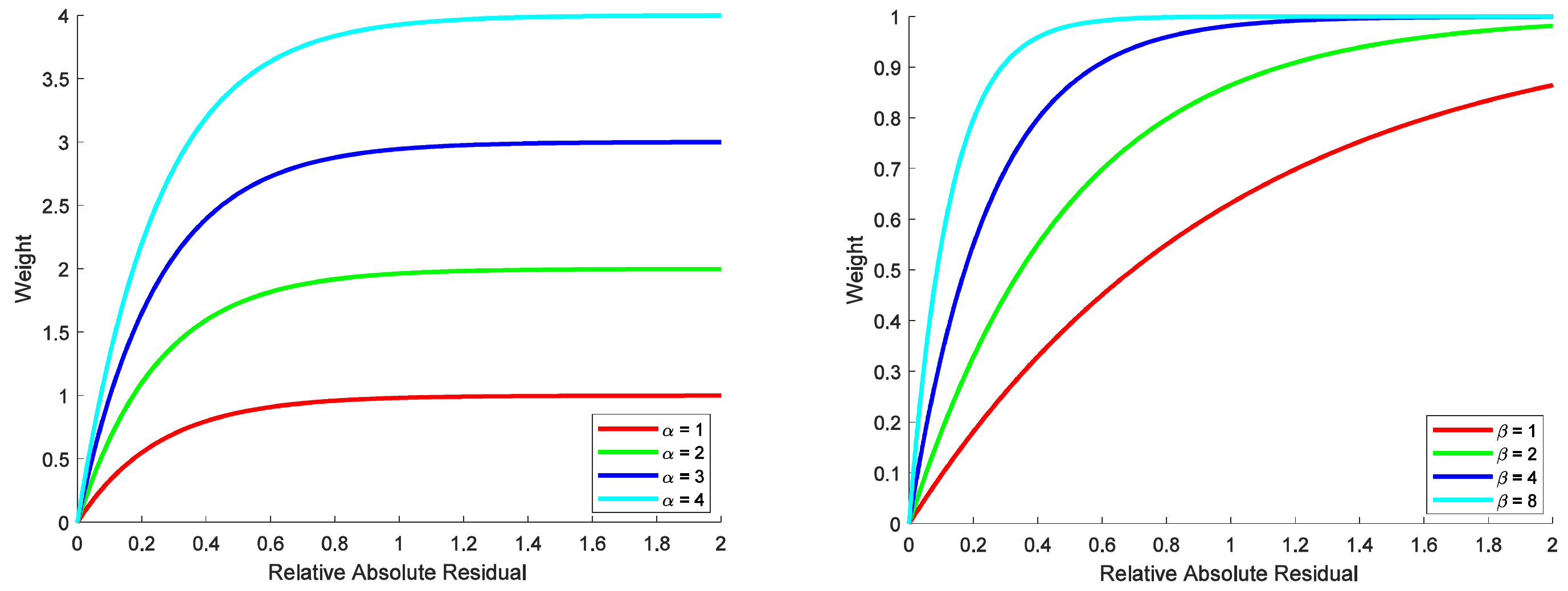}
\caption{Curves of the weight term $\lambda$ in our cost-sensitive Huber loss function. $\beta = 2$ on the left column, and $\alpha = 1$ on the right.}
\label{fig:weight}
\end{figure}

With the structured density maps, we are now ready to define our objective function. As we have pointed out in previous sections, directly learning the inhomogeneous density distribution with the $\ell_2$ loss will cause many problems, e.g., being sensitive to outliers. Therefore, we propose to minimizing the following cost-sensitive Huber loss:

\begin{equation}
L_H = \frac{1}{N}\sum\limits_{n = 1}^1 {{\lambda _n}} \sum\limits_d^D {{H_\delta }({\bf{r}}_n^d)}，
\label{eq:loss}
\end{equation}
where ${\bf{r}}_n^d$ denote the residual on the $d$-th density level, $H_\delta(x)$ is the Huber loss (element-wise) that:
\begin{equation}
{H_\delta }(r) = \left\{ \begin{array}{ll}
  \frac{1}{2}{r^2} & if \, |r| \le \delta ,\\
  \delta (|r| - \frac{1}{2}\delta ), & otherwise;
\end{array} \right.
\label{eq:huberloss}
\end{equation}
and $\lambda_n$ is the weight defined as:
\begin{equation}
\lambda_n = \alpha (1 - {e^{ - \frac{{\beta |{c_n} - c_n^g|}}{{\max (1,c_n^g)}}}}),
\end{equation}
where $\delta$, $\alpha$ and $\beta$ are hyperparameters and $ > 0$, $c_n$ and $c_n^g$ denote the predicted and ground-truth count, respectively. To see how optimizing $L_H$ can obtain a more robust model, one should notice that $\frac{{\partial {H_\partial }}}{{\partial r}} = r$ if $|r| \le \delta$, otherwise $\frac{{\partial {H_\partial }}}{{\partial r}} = \delta sign(r)$, so it clips the gradient to $[-\delta, \delta]$ no matter how large the bias $e$ is. This property reduces the effects of outliers and provides relief from the exploding gradient problem. Furthermore, the weight $\lambda_n$ is introduced to emphasize hard, high-residual examples. The $\frac{{ |{c_n} - c_n^g|}}{{\max (1,c_n^g)}}$ term is the relative absolute residual, and the $max(\cdot,\cdot)$ term is used to prevent from dividing $0$. It is easy to see that $\lambda_n$ is inversely proportional to $\frac{{ |{c_n} - c_n^g|}}{{\max (1,c_n^g)}}$, so that examples with high estimation error can gain more attention in our learning process. The hyperparameter $\alpha$ controls the range of $\lambda_n$ that $\lambda_n \in [0,\alpha]$, and $\beta$ adjusts the slope of curve of $\lambda_n$, as demonstrated in Figure \ref{fig:weight}. We set $\alpha = \beta = 2$ throughout this paper. $\lambda$ and $H_\delta$ together can ensure the gradient stay within a reasonable range, and hence reduce the difficulty of optimization.

\subsection{One Network, Two Goals}

In order to realize the structured density map learning, we propose a simple yet effective Density-Aware Network (DAN) in this section. The intuition behind our design of DAN is straightforward: as to regression, since shallow neural networks can generate rather good density maps, DNNs shall be able to do so as well, due to their higher capacity; as to classification, DNNs are also capable of distinguishing various density levels, e.g., in cutting-edge multi-subnet networks \cite{SwitchCNNCVPR2017,CP-CNN2017}, the very deep VGG-16 \cite{VGG-16} is adopted as the classifier for density levels. Therefore, we believe that a single deep architecture is sufficient for learning the structured density maps. Besides, classifying density levels and learning the density maps are highly related tasks, therefore using a single network can let them share visual cues more effectively.

The proposed DAN is a multi-branch network and its architecture is shown in Figure \ref{fig:architecture}: the ``trunk'' of DAN, namely the shared blocks before splitting into multiple branches, stems from the VGG-16 network. Specifically, we adopt the \emph{Conv1} to \emph{Conv5} blocks in VGG-16, and remain only the first and the second pooling layers in VGG-16, in this way, the downscaling factor $s$ of DAN is $\frac{1}{4}$. We attach $D$ branches to the end of the ``trunk", and each branch consists of two convolutional blocks to generate its corresponding density maps. The specific number of $D$ and the sizes of filters in the branches can be adjusted based on the dataset, and we find $D = 4$ and filters of $5 \times 5$ perform well in most situations. Note that the whole DAN is fully convolutional, therefore it can handle images of arbitrary size.

Besides, in order to prevent the dying ReLU phenomenon, we also change all the activation functions in DAN to the leaky ReLU \cite{rectifier} that $u(x) = max(0.01x,x)$. As to the weight initialization, the weights of filters in \emph{Conv1} to \emph{Conv5} are initialized by the VGG-16 model pre-trained on the ImageNet. For the filters in the branches, as we have analyzed above, high initial value of $\bf{w}$ will cause the exploding gradient problem easily. Therefore, after generating $\bf{w}$ with the Xavier initialization method \cite{Xavier}, we multiply $\bf{w}$ by a small number $\epsilon$, i.e., ${\bf{w}} \leftarrow \epsilon{\bf{w}}$. In our experiments, we find that $\epsilon = 10^{-3}$ can prevent gradient explosion effectively.

Table \ref{tab:mod} summarizes all major modifications in DAN, which can be implemented easily. The architecture of DAN is flexible and can be compatible with most current deep learning frameworks. Unlike other density map based methods, DAN can be trained in a totally end-to-end way, thus makes it a considerable solution for the crowd counting problem.

{\renewcommand{\arraystretch}{1.2}
\begin{table}
\centering
\caption{Summary of major modifications in our method.}
\begin{tabular}{|ll|}
\hline
Architecture:  & VGG-16 {$\rightarrow$} multi-branch networks\\ \hline
Loss function: & $l_2$ loss $\rightarrow$ $L_H$ \\ \hline
Activation function: & ReLU $\rightarrow$ leaky ReLU \\ \hline
Output: & $\bf{g}$ $\rightarrow$ ${\bf{g}}^1,...,{\bf{g}}^D$ \\ \hline
Initialization: & $\bf{w} \rightarrow \epsilon\bf{w}$  \\ \hline
\end{tabular}
\label{tab:mod}
\end{table}}

\begin{table*}[!t]
\centering
\caption{Statistics of datasets for evaluation in our experiments. \textbf{Num} denotes the ratio of training images to test images, \textbf{Range} denotes the range of counts, \textbf{Avg} and \textbf{Std} denote the average and standard deviation of counts, respectively.}
\label{tab:dataset}
\vspace{1em}
\begin{tabularx}{\textwidth}{|Y|Y|Y|Y|Y|}
\hline
\textbf{Dataset}& \textbf{Num} & \textbf{Range} & \textbf{Avg} & \textbf{Std}\\
\hline
UCF \cite{UCF} & 40 : 10 & $[94,4543]$ & 1279.48 & 960.13\\
\hline
WorldExpo\cite{WorldExpo} & 3380 : 600 & $[1,253]$ & 48.29 & 38.22 \\
\hline
ShanghaiTechA\cite{ShanghaiTech} & 300 : 182 & $[33, 3139]$ & 500.29 & 456.56\\
\hline
ShanghaiTechB\cite{ShanghaiTech} & 400 : 316 & $[9, 578]$& 123.59 & 94.52\\
\hline
\end{tabularx}
\end{table*}

\section{Experiments}
This section provides the detailed analysis of our experiments on several dataset \cite{UCF,WorldExpo,ShanghaiTech}. Our  mainly concerns on three aspects: (a) The comparison between the proposed structured inhomogeneous density learning and other state-of-the-art methods (Section \ref{sec:quantitative}), which can show that the proposed method is a simple yet considerable solution for crowd counting in high-density scenes. (b) The ablation study of the proposed Density-Aware Network (DAN) for implementing structured learning on the ShanghaiTech, Part A \cite{ShanghaiTech} (Section \ref{sec:ablation}), which validates our opinions on how the inhomogeneous density distribution affects the learning process. (c) Qualitative results of the proposed method (Section \ref{sec:qualitative}), in which not only the high-quality results, but also those in the worst situations of the proposed methods are demonstrated.

\subsection{Implementation Details}
In order to minimizing the cost-sensitive Huber loss $L_H$, we adopt the Adam optimizer \cite{Adam} with a small learning rate, e.g., $10^{-5}$ to $10^{-7}$. The model will converge in about $20$K iterations. In each iteration, we select a training image and randomly crop a patch of $\frac{1}{4}$ the image size as the training example. We perform data augmentation by horizontally flipping images and adding Gaussian noisy to images.

As to the hyperparameters, the smoothing parameter for soft mapping, $\sigma_v = 2D - 1$; the density thresholds $[S_1,...,S_D]$ depend on the image sizes, and a customary choice is $[3,3^2,...,3^D]$, and the smoothing parameters for Gaussian Filters, $[\sigma_1,...,\sigma_D]$ are set to $0.2 \times [S_1,...,S_D]$; the threshold in Huber loss, $\delta$, is set to $0.2$.

Our implementation of DAN is in MATLAB with the MatConvNet framework \cite{MatConvNet}, source code will be released in the future.

\subsection{Setup}
To validate the effectiveness of our method, we perform extensive experiments on four publicly available datasets, including UCF \cite{UCF}, WorldExpo \cite{WorldExpo}, ShanghaiTech A and B \cite{ShanghaiTech}. The statistics of these datasets are summarized in Table \ref{tab:dataset}. The counts in our experiments cover a wide range from $33$ to $4543$, which can reduce the possibility of over-fitting. For the purpose of fair comparison, following most existing methods, we adopt the mean absolute error (MAE) and the mean square error (MSE) as the metrics for evaluation, which are defined as follows:
\begin{equation}
MAE = \frac{1}{N}\sum\limits_{n = 1}^N {|{c_n} - c_n^g|}, \quad MSE = \sqrt {\frac{1}{N}\sum\limits_{n = 1}^N {{{({c_n} - c_n^g)}^2}}}.
\label{eq:metric}
\end{equation}

\subsection{Quantitative Analysis}
\label{sec:quantitative}

\textbf{ShanghaiTech}: The first dataset we evaluate our method on is the ShanghaiTech dataset \cite{ShanghaiTech} including part A and B. Part A includes $300$ training images and $182$ images with counts ranging from $33$ to $3139$, while Part B includes $400$ training images and $316$ test images with counts ranging from $9$ to $578$. We compare the proposed DAN networks with $7$ deep learning based state-of-the-art methods on this dataset, including Zhang et al. \cite{WorldExpo}, MCNN \cite{ShanghaiTech}, Cascaded-MTL \cite{Cascaded-MTL2017}, Switch-CNN \cite{SwitchCNNCVPR2017}, Huang et al. \cite{BodyCNN2017}, MSCNN \cite{MSCNN} and CP-CNN \cite{CP-CNN2017}.

From the experimental results demonstrated in Table \ref{tab:ShanghaiTech}, we can see that the performance of DAN is significant: it outperforms all state-of-the-art methods by a large margin in Part B, considering both MAE (13.2) and MSE (20.1). As to Part A, DAN only gets the second best performance with respect to MAE. However, note that CP-CNN is a multi-subnet network with higher capacity than DAN, therefore we also compare DAN with the fused network consisting of the global context model (VGG-16) and the density map regressor (MCNN) in CP-CNN (denoted as CP-CNN(G)), and find out that DAN can achieve better accuracy (MAE reduced by $8.1$). Besides, Switch-CNN \cite{SwitchCNNCVPR2017} also employs the VGG-16 architecture to construct one of its subnets, and we can see that the proposed DAN outperforms Switch-CNN on both Part A and Part B. These results indicate that, with our method, a simply end-to-end trained deep network can achieve the state-of-the-art performance.

\begin{table}[!t]
\centering
\caption{Summary of experimental results of the proposed DAN against current methods on the ShanghaiTech dataset \cite{ShanghaiTech}. The best and second methods are marked \First{bold red fonts} and \Second{italic blue fonts}, respectively.}
\label{tab:ShanghaiTech}
\vspace{1em}
\begin{tabular}{|c|c|c|c|c|c|}
\hline
\multicolumn{2}{|c|}{} & \multicolumn{2}{|c|}{\textbf{Part A}} & \multicolumn{2}{|c|}{\textbf{Part B}}\\
\hline
\multicolumn{2}{|c|}{\textbf{Method}} & \textbf{MAE} & \textbf{MSE} & \textbf{MAE} & \textbf{MSE} \\
\hline
\multicolumn{2}{|c|}{Zhang et al. \cite{WorldExpo}} & 181.8 & 277.7 & 32.0 & 49.8 \\
\hline
\multicolumn{2}{|c|}{MCNN \cite{ShanghaiTech}} & 110.2 & 173.2 & 26.4 & 41.3 \\
\hline
\multicolumn{2}{|c|}{Cascaded-MTL \cite{Cascaded-MTL2017}} & 101.3 & 152.4 & 20.0 & 31.1 \\
\hline
\multicolumn{2}{|c|}{Switch-CNN \cite{SwitchCNNCVPR2017}} & 90.4 & 135.0 & 21.6 & 33.4 \\
\hline
\multicolumn{2}{|c|}{Huang et al. \cite{BodyCNN2017}} & - & - & 20.2 & 35.6 \\
\hline
\multicolumn{2}{|c|}{MSCNN \cite{MSCNN}} & 83.8 & \Second{127.4} & \Second{17.7} & 30.2 \\
\hline
\multicolumn{2}{|c|}{CP-CNN (G) \cite{CP-CNN2017}} & 89.9 & 127.9 & - & - \\
\hline
\multicolumn{2}{|c|}{CP-CNN \cite{CP-CNN2017}} & \First{73.6} & \First{106.4} & 20.1 & \Second{30.1} \\
\hline
\multicolumn{2}{|c|}{DAN (ours)} & \Second{81.8} & 134.7 & \First{13.2} & \First{20.1} \\
\hline
\end{tabular}
\label{tab:shanghaiTech}
\end{table}

\begin{table*}[!t]
\centering
\caption{Summary of the MAE of the proposed DAN against existing methods on the WorldExpo dataset \cite{WorldExpo}. The best and second methods are marked \First{bold red fonts} and \Second{italic blue fonts}, respectively.}
\label{tab:WorldExpo}
\vspace{1em}
\begin{tabularx}{\textwidth}{Y|Y|Y|Y|Y|Y|Y|Y|}
\hline
\multicolumn{2}{|c|}{\textbf{Method}} & \textbf{Scene 1} & \textbf{Scene 2} & \textbf{Scene 3} & \textbf{Scene 4} & \textbf{Scene 5} & \textbf{Avg} \\
\hline
\multicolumn{2}{|c|}{Chen et al.\cite{FeatureMining}} & \First{2.1} & 55.9 & \First{9.6} & 11.3 & \First{3.4} & 16.5 \\
\hline
\multicolumn{2}{|c|}{Zhang et al.\cite{WorldExpo}} & 9.8 & \Second{14.1} & 14.3 & 22.2 & 3.7 & 12.9 \\
\hline
\multicolumn{2}{|c|}{MCNN \cite{ShanghaiTech}} & 3.4 & 20.6 & 12.9 & 13.0 & 8.1 & 11.6 \\
\hline
\multicolumn{2}{|c|}{ConvLSTM \cite{convLSTM}} & 7.1 & 15.2 & 15.2 & 13.9 & 3.5 & 10.9 \\
\hline
\multicolumn{2}{|c|}{Huang et al. \cite{BodyCNN2017}} & 4.1 & 21.7 & 11.9 & {11.0} & 3.5 & 10.5 \\
\hline
\multicolumn{2}{|c|}{Switch-CNN with perspective map \cite{SwitchCNNCVPR2017}} & 4.2 & 14.9 & 14.2 & 18.7 & 4.3 & 11.2 \\
\hline
\multicolumn{2}{|c|}{Switch-CNN \cite{SwitchCNNCVPR2017}} & 4.4 & 15.7 & \Second{10.0} & {11.0} & 5.9 & 9.4 \\
\hline
\multicolumn{2}{|c|}{CP-CNN \cite{CP-CNN2017}} & \Second{2.9} & 14.7 & 10.5 & \First{10.4} & 5.8 & \First{8.86} \\
\hline
\multicolumn{2}{|c|}{DAN (ours)} &  4.1 & \First{11.1} &  10.7 & 16.2 & 5.0 & 9.4 \\
\hline
\end{tabularx}
\end{table*}

\textbf{WorldExpo}: The second dataset for evaluation is WorldExpo \cite{WorldExpo}, which contains test images from $5$ scenes. We compare DAN with $7$ state-of-the-art methods on this dataset, including Chen et al.\cite{FeatureMining}, Zhang et al.\cite{WorldExpo}, MCNN \cite{ShanghaiTech}, ConvLSTM \cite{convLSTM}, Huang et al. \cite{BodyCNN2017}, Switch-CNN \cite{SwitchCNNCVPR2017} and CP-CNN \cite{CP-CNN2017}.

As demonstrated in Table \ref{tab:WorldExpo}, the performance of all methods are very close (from $8.86$ to $16.5$). The major reasons behind this are dual: on the one hand, counts in this dataset range from $1$ to $253$, with the smallest standard deviation value ($38.22$) among all the datasets, therefore it is hard to distinguish the performance of different methods based on the MAE metric; on the other hand, this dataset provides the ground-truth perspective maps for generating training templates, which obviously reduces the effects of outliers. Nevertheless, the proposed DAN still gets the lowest MAE in Scene 5, and the second lowest MAE considering the average MAE across all scenes. Furthermore, we do not use the ground-truth perspective maps in the training process of DAN, which indicates our method is robust to variant perspectives.

\textbf{UCF}: The UCF \cite{UCF} is the most difficult dataset, because it has only $40$ images for training and has the highest standard deviation value ($960.13$). Similar to other methods, we perform 5-fold cross validation on this dataset. We compare DAN with $12$ state-of-the-art methods on UCF, including Lempitsky et al. \cite{Lempitsky}, Idrees et al. \cite{UCF}, Zhang et al. \cite{WorldExpo}, MCNN \cite{ShanghaiTech}, Huang et al. \cite{BodyCNN2017}, CCNN \cite{HyderNet}, MSCNN \cite{MSCNN}, CNN Boosted \cite{boostedCNN}, Hydra 2s \cite{HyderNet}, Cascaded-MTL \cite{Cascaded-MTL2017}, Switch-CNN \cite{SwitchCNNCVPR2017} and CP-CNN \cite{CP-CNN2017}.

The experimental results are shown in Table \ref{tab:UCF}. From this result we can see that DAN can achieve rather high accuracy, its MAE is $309.6$, which is between that of Switch-CNN ($318.1$) and CP-CNN ($295.8$). This again validates our opinion that a single deep network with proper learning settings can reduce the effects of outliers significantly.

\begin{table}[!t]
\centering
\caption{Summary of experimental results of the proposed DAN against current methods on the UCF dataset \cite{UCF}. The best and second methods are marked \First{bold red fonts} and \Second{italic blue fonts}, respectively.}
\vspace{1em}
\begin{tabular}{|c|c|c|}
\hline
\textbf{Method} & \textbf{MAE} & \textbf{MSE} \\
\hline

Lempitsky et al. \cite{Lempitsky} & 493.4 & 487.1 \\ \hline

Idrees et al. \cite{UCF} & 419.5 & 541.6 \\ \hline

Zhang et al. \cite{WorldExpo} & 467.0 & 498.5 \\ \hline

MCNN \cite{ShanghaiTech} & 377.6 & 425.2 \\ \hline

Huang et al. \cite{BodyCNN2017} & 409.5 & 563.7 \\ \hline

CCNN \cite{HyderNet} & 488.67 & 646.68 \\ \hline

MSCNN \cite{MSCNN} & 363.7 & 468.4 \\ \hline

CNN Boosted \cite{boostedCNN} & 364.4 & - \\ \hline

Hydra 2s \cite{HyderNet} & 333.73 & 425.26 \\ \hline

Cascaded-MTL \cite{Cascaded-MTL2017} & 322.8 & 397.9 \\ \hline

Switch-CNN \cite{SwitchCNNCVPR2017} & 318.1 & 439.2 \\ \hline

CP-CNN \cite{CP-CNN2017} & \Second{295.8} & \Second{320.9} \\ \hline

ConvLSTM \cite{convLSTM} & \First{284.5} & \First{297.1} \\ \hline

DAN (ours) & 309.6 & 402.64 \\ \hline
\end{tabular}
\label{tab:UCF}
\end{table}

\subsection{Ablation Study}
\label{sec:ablation}
Remember that the major conclusion of this paper is that the inhomogeneous density distribution problem will result in the bias to outliers and other difficulties in the learning process. To validate this, we conduct an ablation study on the ShanghaiTech dataset with Part A. We compare the performance of DAN with different settings: training the VGG-16 part in DAN with $\ell_2$-loss (denoted as VGG-16), DAN adopting Huber loss and only $1$ branch (denoted as DAN-1-H), and DAN with different numbers of branches (denoted as DAN-1 to DAN-4). All the variants of DAN are trained with the same settings.

\begin{table}[!h]
\centering
\caption{Summary of experimental results of our ablation study on the ShanghaiTech dataset, part A \cite{ShanghaiTech}. VGG-16 denotes directly training the VGG-16 part in DAN with $\ell_2$-loss; DAN-* denotes DAN with * branches, and DAN-1-H denotes DAN without the weighting term in the cost-sensitive Huber loss. Due to the gradient explosion problem, VGG-16 cannot be trained and its result is omitted.}
\vspace{1em}
\begin{tabular}{|c|c|c|}
\hline
\textbf{Method} & Density Thresholds & \textbf{MAE} \\
\hline
VGG-16 & $ - $ & - \\ \hline
DAN-1-H & $[49]$ & 87.0 \\ \hline
DAN-1 & $[49]$ & 84.3 \\ \hline
DAN-2 & $[25, 81]$ & 83.0 \\ \hline
DAN-3 & $[9, 49, 121]$ & 83.8 \\ \hline
DAN-4 & $[9,25,49,81]$ & 81.8 \\ \hline
\end{tabular}
\label{tab:albation}
\end{table}

The experimental results are shown in Table \ref{tab:albation}. Firstly, we have tried to train the VGG-16 with $\ell_2$ loss several times, but unfortunately, all end up with the exploded gradients. This confirms our opinion that outliers can cause difficulties in the learning process. With the Huber loss and other modifications in our network, we succeed in learning the deep networks (DAN-1-H), which also outperforms most current networks. Secondly, our weighting strategy in the loss function can increase the performance significantly, as the performance of DAN-1 is higher than that of DAN-1-H. Lastly, we test DAN with different number of branches, of which the density thresholds are set to be distributed as equally as possible, e.g., selected from $[3,5,7,9,11]^2$. We find that with more branches, the performance of DAN is getting better. This can be understood as with more branches, the differences between training templates and ideal templates are smaller, and more local models are ensembled, and consequently the networks are more robust.

\begin{figure*}[!t]
\centering
\includegraphics[width = 1\textwidth]{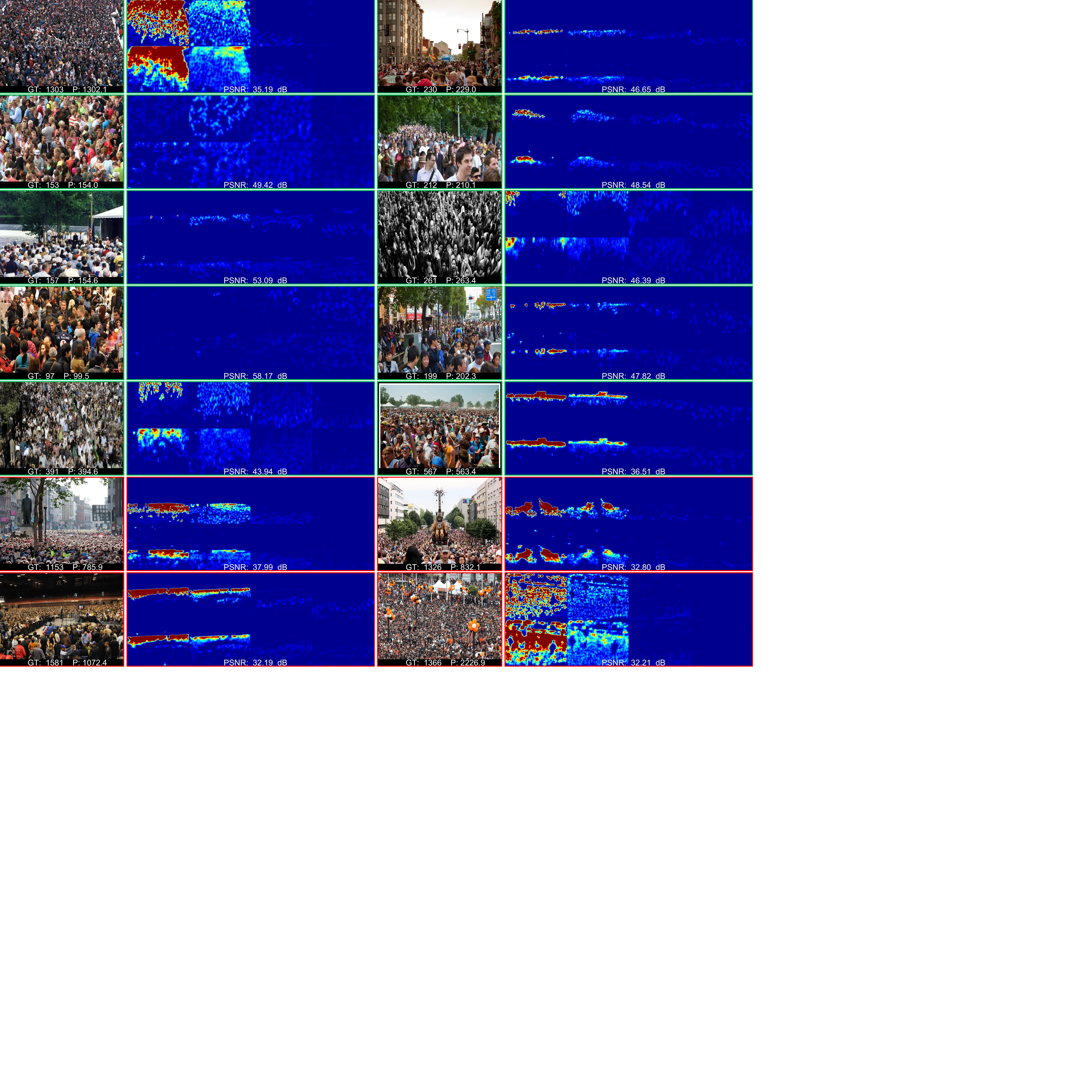}
\caption{Qualitative results of the proposed method on the ShanghaiTech dataset \cite{ShanghaiTech}, Part A. For each image, we demonstrate its ground-truth structured density map of $4$ levels on the top, while the predicted one on the bottom. Brighter color indicates higher density. The ground-truth (GT), predicted count (P), and the Peak Signal to Noise Ratio (PSNR) of our method are marked on the bottom of each image. Successful cases are marked with green while failure cases are marked with red. Our method can generate high-accuracy results.}
\label{fig:qualityA}
\end{figure*}

\begin{figure*}[!t]
\centering
\includegraphics[width = 1\textwidth]{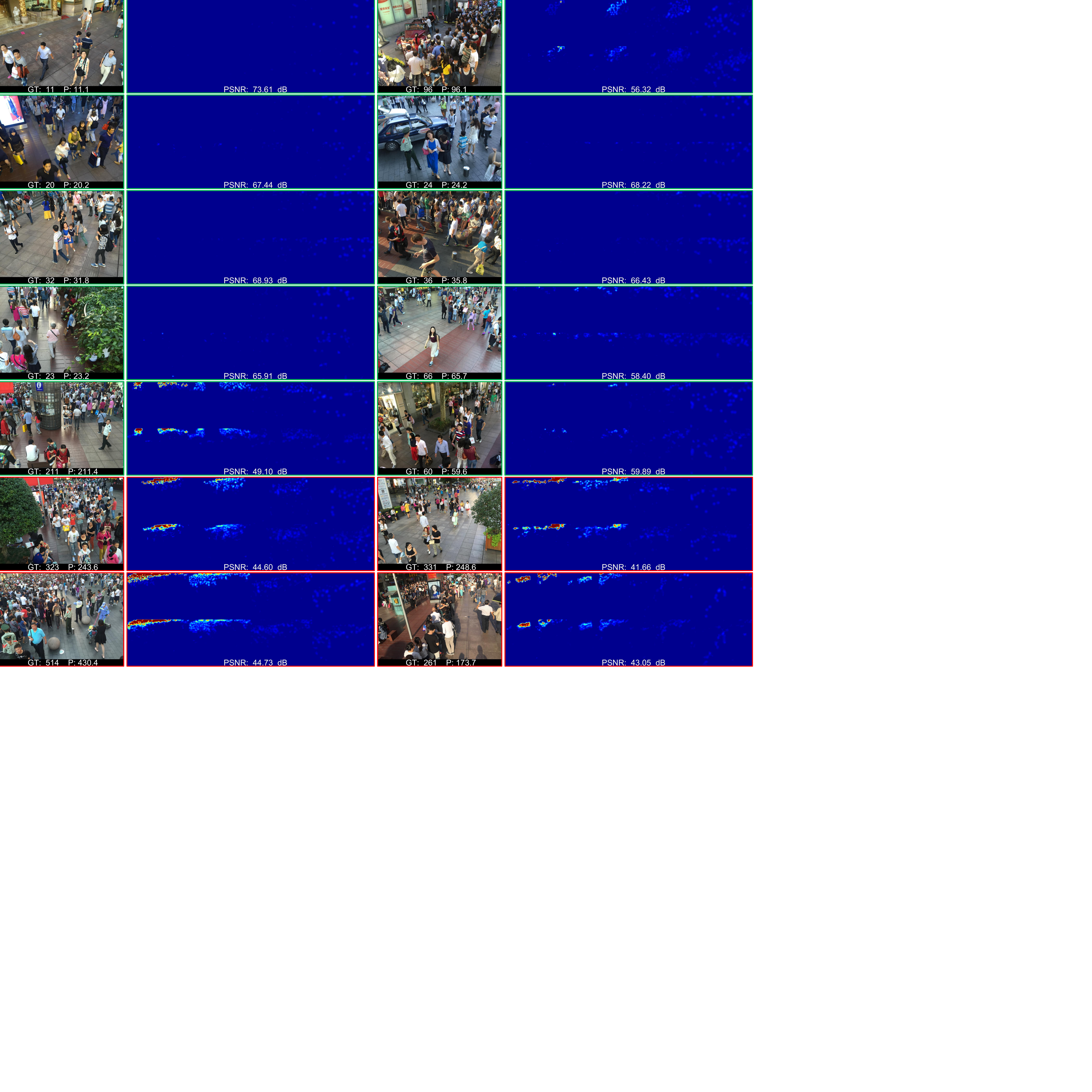}
\caption{Qualitative results of the proposed method on the ShanghaiTech dataset \cite{ShanghaiTech}, Part B. For each image, we demonstrate its ground-truth structured density map of $4$ levels on the top, while the predicted one on the bottom. Brighter color indicates higher density. The ground-truth (GT), predicted count (P), and the Peak Signal to Noise Ratio (PSNR) of our method are marked on the bottom of each image. Successful cases are marked with green while failure cases are marked with red. Our method can generate high-accuracy results.}
\label{fig:qualityB}
\end{figure*}

\subsection{Qualitative Analysis}
\label{sec:qualitative}
We demonstrate representative examples of our experimental results on the ShanghaiTech for qualitative analysis in this section. Figure \ref{fig:qualityA} and \ref{fig:qualityB} shows $14$ images with their ground-truth structured density maps and predicted density maps on Part A and B, respectively. As in \cite{CP-CNN2017}, we also calculate the Peak Signal to Noise Ratio (PSNR) as the metric to evaluate the quality of our predicted density maps.

From these results we can see that the proposed method perform well in most situations, not matter for extremely high-density scenes (Part A) or less crowd scenes (Part B). Besides, most PSNR values of our predicted density maps are above $40$ dB, which indicates DAN can generate high-quality density maps. More importantly, notice that our predicted structured density maps are divided into $4$ levels clearly, therefore validates our opinion that DAN can accomplish the regression task and the classification task at the same time.

Yet there are a few failure cases for the proposed method. To analyze this, we demonstrate the top-4 images ranked by the MAE in Figure \ref{fig:qualityA} and \ref{fig:qualityB}. We notice that a common characteristic of these images is that in the extremely high-density areas of these images, due to the restricted resolutions, even human cannot precisely distinguish the targets. Especially for images in Part A, sizes of targets are even less than $10$ pixels, e.g., the last row in Figure \ref{fig:qualityA}. Therefore, for these images, the resolution of our network is not high enough to handle them. Fortunately, these images are rare in all datasets, hence we still can consider DAN as a practical solution for the crowd counting problem.

\section{Conclusion}

In this paper, we focus on the crowd counting problem in extremely high-density scene images and take in-depth analysis into the phenomena that existing deep learning-based methods do not work well while deeper networks are employed. Our comprehensive study reals that it can be heavily attributed to the inhomogeneous density distribution problem, and a feasible solution is provided by extending the density map from 2D to 3D, with a extra dimension implicitly indicating the density level. Based on this, we also present a single Density-Aware Network that is simple and easy to train. Extensive experiments demonstrate that it achieves the state-of-art performance on several challenging datasets.

{\small
\bibliographystyle{ieee}
\bibliography{egbib}
}

\end{document}